\documentclass[letterpaper, 10 pt, conference]{ieeeconf}  

\IEEEoverridecommandlockouts                              

\usepackage{graphicx}
\usepackage{epstopdf}
\usepackage{subcaption}
\usepackage{color}
\usepackage{multirow}
\usepackage{adjustbox}
\usepackage{booktabs}
\usepackage[table,xcdraw]{xcolor}
\usepackage{caption}
\usepackage{amsmath}
\usepackage{url}
\usepackage{array} 
\usepackage{graphicx}
\usepackage{amsmath}
\usepackage{amssymb}
\usepackage{booktabs}
\usepackage{cite}
\usepackage{bbding}

\definecolor{others}{rgb}{0, 0, 0}
\definecolor{barrier}{rgb}{1, 0.47058824, 0.19607843}
\definecolor{bicycle}{rgb}{1, 0.75294118, 0.79607843}
\definecolor{bus}{rgb}{1, 1, 0.0}
\definecolor{car}{rgb}{0.0, 0.58823529, 0.96078431}
\definecolor{construction}{rgb}{0, 1, 1}
\definecolor{motorcycle}{rgb}{1, 0.49803922, 0}
\definecolor{pedestrian}{rgb}{1, 0, 0}
\definecolor{cone}{rgb}{1, 0.94117647, 0.58823529}
\definecolor{trailer}{rgb}{0.52941176, 0.23529412, 0}
\definecolor{truck}{rgb}{0.62745098, 0.1254902, 0.94117647}

\definecolor{driveable}{rgb}{1, 0, 1}
\definecolor{flat}{rgb}{0.54509804,0.5372549,0.5372549}
\definecolor{sidewalk}{rgb}{0.29411765,0,0.29411765}
\definecolor{terrain}{rgb}{0.58823529,0.94117647,0.31372549}
\definecolor{manmade}{rgb}{0.90196078,0.90196078,0.98039216}
\definecolor{vegetation}{rgb}{0,0.68627451,0}
\definecolor{ego_vehicle}{rgb}{0,0,0}

\title{\LARGE \bf
OccCylindrical: Multi-Modal Fusion with Cylindrical Representation for 3D Semantic Occupancy Prediction
}

\author{Zhenxing Ming, Julie Stephany Berrio, Mao Shan, Yaoqi Huang, \\ Hongyu Lyu, Nguyen Hoang Khoi Tran, Tzu-Yun Tseng, and Stewart Worrall
\thanks{The authors are with the Australian Centre for Robotics (ACFR) at the University of Sydney (NSW, Australia). E-mails: {\tt\small{\{d.ming, j.berrio, m.shan, y.huang, h.lyu, n.tran, t.tseng, s.worrall}\}@acfr.usyd.edu.au}.}%
}

\begin{document}

\maketitle
\thispagestyle{empty}
\pagestyle{empty}

\begin{abstract}
The safe operation of autonomous vehicles (AVs) is highly dependent on their understanding of the surroundings. For this, the task of 3D semantic occupancy prediction divides the space around the sensors into voxels, and labels each voxel with both occupancy and semantic information.
Recent perception models have used multisensor fusion to perform this task. 
However, existing multisensor fusion-based approaches focus mainly on using sensor information in the Cartesian coordinate system. This ignores the distribution of the sensor readings, leading to a loss of fine-grained details and performance degradation. In this paper, we propose OccCylindrical that merges and refines the different modality features under cylindrical coordinates. Our method preserves more fine-grained geometry detail that leads to better performance.  Extensive experiments conducted on the nuScenes dataset, including challenging rainy and nighttime scenarios, confirm our approach's effectiveness and state-of-the-art performance. The code will be available at: \url{https://github.com/DanielMing123/OccCylindrical}

\end{abstract}


\section{INTRODUCTION}
Understanding and modelling the surrounding scene is crucial for the safety of autonomous vehicles (AVs). AVs are usually equipped with multiple sensors, such as surround-view cameras, surround-view radars, and high-density LiDARs, to provide better perception capabilities. The introduction of a 3D semantic occupancy \cite{tpvformer,occformer,surroundocc} representation further enhances the perception capability of AVs. Multisensor fusion based approaches take surround view cameras, LiDAR, and radar information as input to predict 3D semantic occupancy. This is becoming a popular research direction due to the complementary information between the different modality sensors and robustness against challenging lighting and weather conditions.

\begin{figure}[t]
\centering
{\includegraphics[width=0.95\columnwidth]{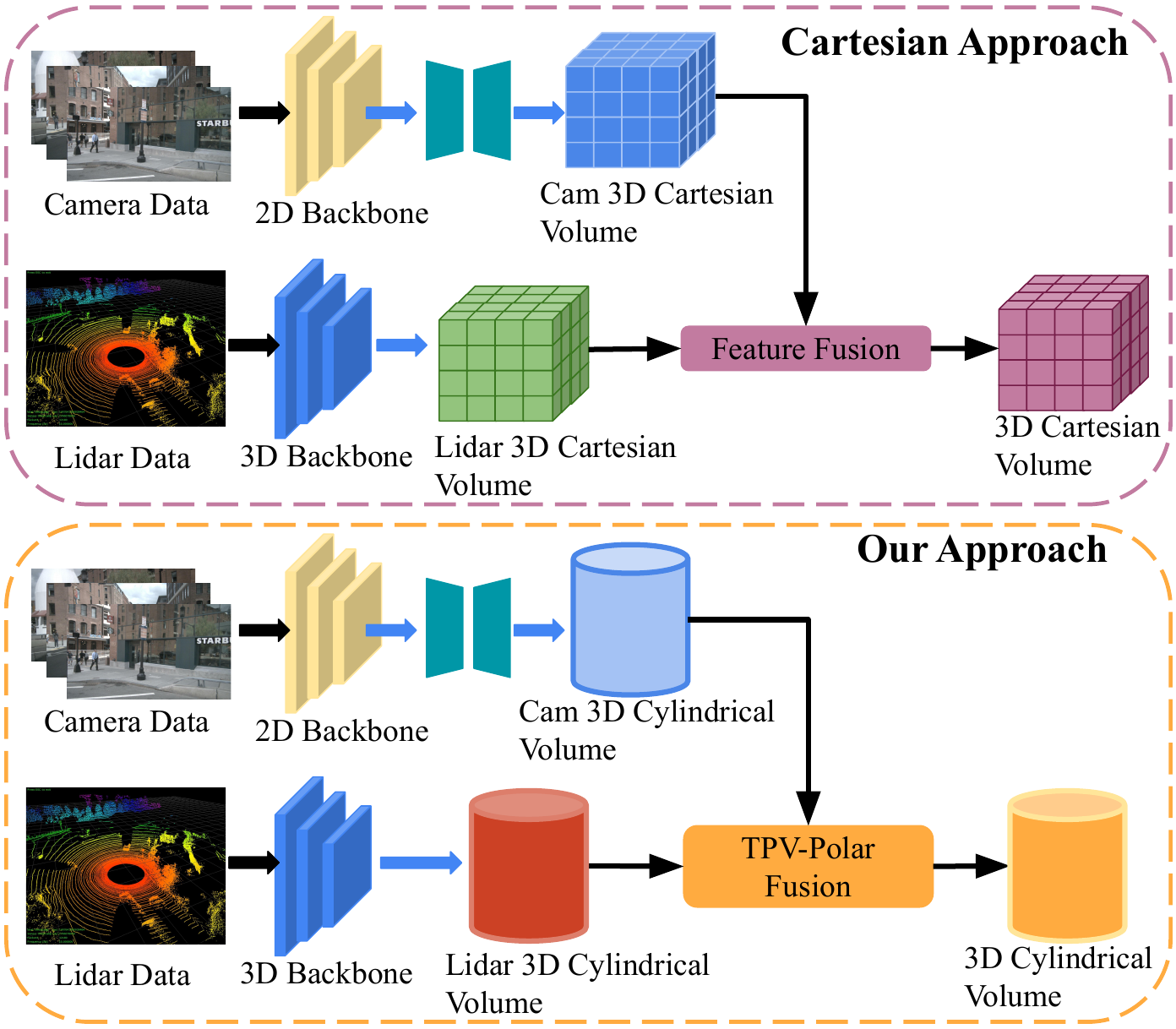}}
\caption{\small Pipeline for two approaches: Cartesian approach (top) and our approach (bottom). 
In contrast to the current multi-sensor fusion frameworks that process 3D feature volumes under Cartesian coordinates, we perform 3D semantic occupancy prediction through feature fusion across two distinct modalities under cylindrical coordinates. 
This takes into account the distribution of the sensor readings, preserving more fine-grained geometry information.
}\label{teaser}
\end{figure}

The existing methods focus mainly on information interaction and fusion among different sensor modalities. By overlooking the information distribution characteristics of each sensor, these methods fail to preserve fine-grained details effectively, which, in turn, limit their overall model performance. For example, the 3D point cloud generated by LiDAR is dense in the nearby region and increasingly sparse at greater distances. Similarly, due to the perspective effect in the images, objects closer to the sensor appear larger, while distant objects appear smaller.
To address the aforementioned limitations and preserve the fine-grained information as much as possible, we propose OccCylindrical (Fig.\ref{teaser} bottom). Compared to the existing approaches \cite{inverse++,sparseocc,geocc,2dpass,occ3d,bevfusion,fusionocc,occfusion} (Fig. \ref{teaser} top) that conduct feature refinement and fusion under Cartesian coordinates, we carry out the above operations under cylindrical coordinates to match the 3D point distribution of the LiDAR point cloud. We generate a pseudo-3D point cloud from the camera branch and convert it to cylindrical coordinates to perform feature refinement and fusion. The intermediate 3D cylindrical volumes are further compressed along each axis, resulting in a set of tri-perspective view (TPV) polar planes, followed by a shared encoder-decoder to refine the features further. Through comparisons with other state-of-the-art (SOTA) algorithms on the nuScenes dataset, including challenging rainy and nighttime scenarios, we demonstrate our method's effectiveness and robustness.

The main contributions of this paper are summarised as follows:
\begin{itemize}
    \item We propose OccCylindrical, a novel 3D semantic occupancy prediction framework based on multisensor fusion that uses cylindrical coordinates to preserve better fine-grained geometry information for each sensor modality.
    \item We introduce a novel depth estimation module that enhances the visual feature context with depth information and position embedding by generating a high-quality pseudo-3D point cloud from surround-view cameras.
    \item We propose the TPV-Polar-Fusion module, which performs spatial group pooling and attention-based feature fusion under cylindrical coordinates to preserve fine-grained geometry information from each modality.
    \item We compare our approach with other SOTA algorithms in the 3D semantic occupancy prediction task to demonstrate the effectiveness of our method.
\end{itemize}

The remainder of this paper is structured as follows: Section \ref{literature} provides an overview of related research and identifies the key differences between this study and previous publications. Section \ref{model} outlines the general framework of OccCylindrical and offers a detailed explanation of the implementation of each module. Section \ref{simulation} presents the results of our experiments. Finally, Section \ref{conclusion} provides the conclusion of our work.
\section{Related Work}\label{literature}
\subsection{Camera-Radar Fusion-based 3D semantic occupancy prediction}
In \cite{hydra}, the authors perform early-stage feature fusion for surround-view cameras and radars through a proposed height association transformer module. This method first performs early-stage feature fusion by projecting the sparse radar point cloud onto surround-view visual features, and then applies a radar-weighted backward projection for middle-stage feature fusion.
Their approach achieved remarkable performance on both 3D semantic occupancy prediction and 3D object detection. Feature processing is performed in Cartesian coordinates using the Bird's Eye View (BEV) format to retain relevant spatial information. In contrast, our approach uses cylindrical coordinates, which better align with the way point clouds are generated by rotational LiDARs. This alignment preserves geometric consistency, enabling more effective feature fusion and intermediate feature refinement, leading to minimal information loss and improved overall performance. 

\subsection{Camera-LiDAR Fusion-based 3D semantic occupancy prediction}
In \cite{occfusion}, the authors employ VoxNet as the 3D backbone to extract features from LiDAR and radar data, producing an initial volume of 3D features. Subsequently, a view transformation method introduced in \cite{inversematrixvt3d} is applied to generate an additional 3D feature volume. These volumes are then combined using a feature fusion module for intermediate-stage fusion, resulting in the final merged 3D feature representation.

Similarly, in \cite{fusionocc}, the authors leverage the view transformation proposed in \cite{LSS} to generate a pseudo-3D point cloud under Cartesian coordinates from surround-view cameras. Then, the proposed depth encoder module refines the pseudo-3D point cloud. Applying voxelization to the LiDAR-generated real 3D point cloud and the camera-generated pseudo-3D point cloud produces two distinct 3D feature volumes, each representing a different modality. Finally, a cross-modality fusion module is introduced to integrate the two 3D feature volumes, resulting in the final fused 3D representation. Although both methods demonstrate strong performance on benchmark datasets, they rely on voxelization in Cartesian coordinates and fail to account for the non-uniform distribution of 3D point clouds in space—particularly the density variation caused by sensor characteristics. This results in a large number of empty voxel grids, especially in distant regions, and leads to the loss of fine-grained geometric details, ultimately limiting the effectiveness of feature representation in those areas.

In contrast, our approach processes the pseudo and real 3D point clouds, which come from surround-view cameras and lidar, respectively, under cylindrical coordinates for better representation and to preserve more fine-grained geometry information that leads to better performance.

\section{OccCylindrical}\label{model}
\begin{figure*}[h]
    \vspace{2mm}
     \centering
     \begin{subfigure}[]{0.99\textwidth}
         \centering
         \includegraphics[width=\textwidth]{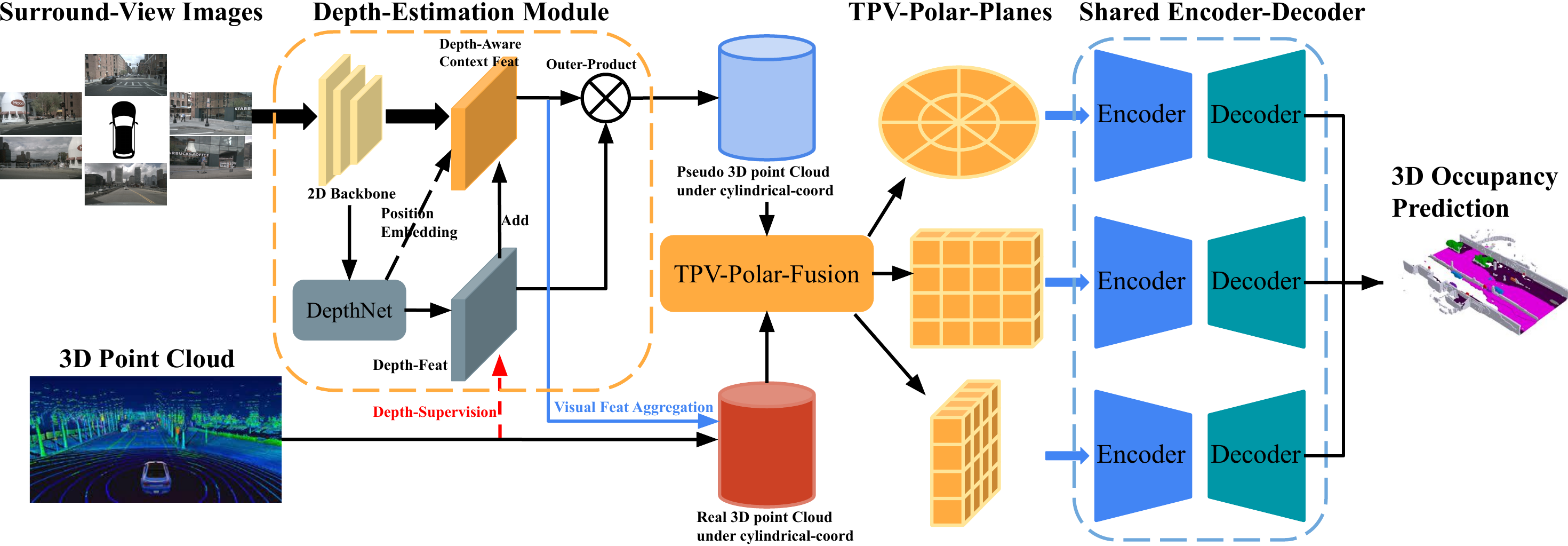}
     \end{subfigure}
        \caption{\small \textbf{Overall architecture of OccCylindrical.} The surround-view images are initially processed through the 2D backbone to extract visual features. Subsequently, a DepthNet is employed to generate a depth distribution feature based on these visual features. The depth distribution feature is supervised using depth information from the LiDAR point cloud. Meanwhile, the predefined depth-distribution coordinate, used as a positional embedding, along with the depth feature, is fused back into the visual feature, resulting in a Depth-Aware context feature. The outer product operation is applied on the depth distribution feature and depth-aware context feature, resulting in a pseudo-3D point cloud. The TPV-Polar-Fusion module takes the pseudo-3D point cloud and the LiDAR point cloud as input to do the feature-level fusion and outputs three TPV-Polar planes. The shared encoder-decoder structure further refines the TPV-Polar planes and outputs TPV-Polar planes to the prediction head for the 3D semantic occupancy prediction.}
        \label{OccCylindrical}
\end{figure*}

\subsection{Problem Statement}
This paper aims to generate a dense 3D semantic occupancy grid of the environment surrounding an autonomous vehicle using data from surround view cameras and LiDAR sensors onboard. Therefore, the problem can be formulated as:
\begin{equation}
    Occ = M(Cam^{1},Cam^{2}...,Cam^{N},LiDAR)
\end{equation}
where $M$ represents the fusion framework integrating camera and LiDAR information together to do the 3D semantic occupancy prediction. The final 3D semantic occupancy prediction result is represented by $Occ \in R^{X \times Y \times Z}$, where each grid is assigned a semantic property ranging from 0 to 17. In our case, a class value of 0 corresponds to an empty grid.

\subsection{Overall Architecture}
The overall architecture is shown in Figure \ref{OccCylindrical}. Initially, given the surround view images $Img=\left \{ img^{1}, img^{2}, img^{3},....,img^{N} \right \}$, and the dense 3D point cloud from LiDAR $P^{LiDAR}$, we apply a 2D backbone (e.g. ResNet50) to obtain a single-scale visual feature $V\in R^{N\times C\times H\times W}$. Then, the DepthNet proposed in \cite{bevdepth} is applied to the visual feature $V$, resulting in the depth distribution feature $V_{depth}\in R^{N\times D \times H \times W}$ and its associate depth-bin coordinates $V_{depth\_coord}\in R^{N\times D \times H \times W \times 3}$. Meanwhile, the LiDAR 3D point cloud from LiDAR is used to perform depth supervision with respect to the depth distribution feature $V_{depth}$. The depth distribution feature, along with each depth bin coordinate, which serves as a position embedding, is added to the visual feature $V$, resulting in the depth-aware context feature $V_{context}\in R^{N \times C \times H \times W}$. An outer product operation creates a pseudo-3D point cloud $P^{Cam}\in R^{x\times y\times z\times C}$ from a depth-aware context feature and the depth distribution feature. The pseudo-3D point cloud is converted into a cylindrical coordinate $P^{Cam}_{Cylin}\in R^{r\times \theta \times z\times C} $ for better representation. 

The 3D point cloud from LiDAR $P_{raw}^{LiDAR}$ first goes through an MLP layer to extract geometric features $P^{LiDAR}_{geo}\in R^{x\times y\times z\times C1}$, and then each 3D point is projected onto the depth-aware context feature to aggregate the semantic information, resulting in $P^{LiDAR}_{sem}\in R^{x\times y\times z\times C2}$. Then $P^{LiDAR}_{geo}$ and $P^{LiDAR}_{sem}$ are concatenated through feature channel dimension and input to another MLP layer to perform feature filtering, resulting in the final $P^{LiDAR}\in R^{x\times y\times z\times C}$. This process is described as follows: 
\begin{equation}
    P_{geo}^{LiDAR}=MLP(P_{raw}^{LiDAR} ) 
\end{equation}
\begin{equation}
    P_{sem}^{LiDAR}=S(P_{raw}^{LiDAR}, V_{context}) 
\end{equation}
\begin{equation}
    P^{LiDAR}=MLP(Cat(P_{geo}^{LiDAR},P_{sem}^{LiDAR})) 
\end{equation}
where $S$ refers to the projection and sampling operation. Then, $P^{LiDAR}$ is converted to the cylindrical coordinate $P^{LiDAR}_{Cylin}\in R^{r\times \theta \times z\times C} $ for better representation. The TPV-Polar-Fusion module takes $P^{Cam}_{Cylin}$ and $P^{LiDAR}_{Cylin}$ as input to generate triperspective view (TPV) polar planes. These planes are further refined by a shared 2D encoder-decoder, resulting in multiscale TPV-Polar planes. Lastly, the multiscale TPV-polar planes are fed to the task head for 3D semantic occupancy prediction.

\subsection{Depth-Estimation Module}
In the depth estimation module, surround view images are first processed by a 2D backbone, resulting in the single-scale visual feature $V$, which is $1/16$ of the original image resolution. Following the procedure proposed in \cite{LSS,bevdet,bevdepth}, we develop the depthnet with LiDAR Cartesian coordination. Within depthnet, we predefined a set of depth bins along the ray of each visual feature grid, and the interval $d$ between each depth bin is set as $d=1$ meter. The farthest depth bin along the ray is located 50 meters from the ray's origin. 
Each depth bin corresponds to a 3D point in Cartesian coordinates and represents the probability of this volume being occupied. The depth feature is supervised using the LiDAR-generated 3D point cloud, enabling it to capture the geometric structure of the 3D scene more effectively. Furthermore, inspired by PETR \cite{petr}, we convert each Cartesian depth bin coordinate into the cylindrical coordinate and add it back as position embedding together with depth features to the original visual feature $V$, resulting in the depth-aware context feature $V_{context}\in  R^{N\times C\times H\times W} $. This process is described as follows:
\begin{equation}
    V_{context} = V + CNN_{1}(V_{depth}) + CNN_{2}(V_{coord})
\end{equation}
where $V_{coord}$ refers to the 3D cylindrical coordinate of all depth bins and $CNN_{1}$ and $CNN_{2}$ are Conv2D-based small network to project $V_{depth}$ and $V_{coord}$ to the same feature channel dimension as $V$. Finally, the outer product operation is applied on $V_{depth}$ and $V_{context}$, followed by the Cartesian to cylindrical coordinate conversion, resulting in the final pseudo-3D point cloud in cylindrical coordinates.

\subsection{TPV-Polar-Fusion}
Given real and pseudo-3D point clouds in cylindrical coordinates $P^{Cam}_{Cylin}$ and $P^{LiDAR}_{Cylin}$, the TPV-Polar-Fusion module first applies a cylindrical partitioning to the two 3D point clouds, transforming them into cylindrical volumes with a resolution of $[R, D, Z]$. Since each cylindrical voxel may contain multiple points, a MaxPooling operation is applied within each voxel to aggregate the features, resulting in the final cylindrical voxel representation as follows:
\begin{equation}
    V_{Cylin}^{Cam} = MaxPool(P_{Cylin}^{Cam},p_{coord}^{Cam})
\end{equation}
\begin{equation}
    V_{Cylin}^{LiDAR} = MaxPool(P_{Cylin}^{LiDAR},p_{coord}^{LiDAR})
\end{equation}
To obtain 2D TPV polar planes, we further compress the cylindrical volume $V_{Cylin}\in R^{R\times D\times Z\times C}$ along three axes. Following the study in \cite{pointocc}, we adopt spatial group pooling instead of MaxPooling to better preserve the fine-grained geometry details. For the spatial group pooling, we split the cylindrical volume $V_{Cylin}$ along the pooling axes $R$, $D$, and $Z$ into $M$ groups and perform MaxPooling in each group, respectively. Then, the resulting $M$ group features are concatenated along the channel dimension and input into an MLP layer to perform the feature filtering. These processes are described as follows:
\begin{equation}
    F_{RD} = MLP(Cat(\left \{V^{R\times D\times C}_{i} \right \} _{i=1}^{M} ))
\end{equation}
\begin{equation}
    F_{DZ} = MLP(Cat(\left \{V^{D\times Z\times C}_{i} \right \} _{i=1}^{M} ))
\end{equation}
\begin{equation}
    F_{ZR} = MLP(Cat(\left \{V^{Z\times R\times C}_{i} \right \} _{i=1}^{M} ))
\end{equation}
where
\begin{equation}
    V^{R\times D\times C}_{i} = MaxPool_{\left \{ Z,i \right \} }(V^{R\times D\times Z_{i}})
\end{equation}
\begin{equation}
    V^{D\times Z\times C}_{i} = MaxPool_{\left \{ R,i \right \} }(V^{D\times Z\times R_{i}})
\end{equation}
\begin{equation}
    V^{Z\times R\times C}_{i} = MaxPool_{\left \{ D,i \right \} }(V^{Z\times R\times D_{i}})
\end{equation}
and $MaxPool_{\left \{ Z/R/D,i \right \} }$ refers to apply MaxPooling operation along the specific axis on $i$-th group of cylindrical volume. The $V_{Cylin}^{Cam}$ and $V_{Cylin}^{LiDAR}$ after being processed by spatial group pooling resulting in two sets of TPV-Polar planes $\left \{ F_{RD}^{Cam}, F_{DZ}^{Cam}, F_{ZR}^{Cam} \right \}$ and $\left \{ F_{RD}^{LiDAR}, F_{DZ}^{LiDAR}, F_{ZR}^{LiDAR} \right \}$. Then, the 2D dynamic fusion module proposed in \cite{occfusion} is applied to each 2D plane from two sets respectively, resulting in the final fused TPV-Polar planes $\left \{ F_{RD}^{fuse}, F_{DZ}^{fuse}, F_{ZR}^{fuse} \right \} $.

\subsection{Shared Encoder-Decoder}
To enhance the fused TPV-Polar planes, we adopt the swin-transformer \cite{swintrans} as the encoder to refine the feature, followed by an FPN as the decoder to generate multi-scale TPV-Polar planes. Specifically, three TPV-Polar planes share the same encoder and decoder, and the output of the decoder is refined TPV-Polar planes in four scales $F_{refined} =\left \{ \left \{ F_{RD}^{l}, F_{DZ}^{l}, F_{ZR}^{l} \right \}  \right \}_{l=1}^{L=4}$. These processes are described as:
\begin{equation}
    \left \{ F_{RD}^{l}  \right \}_{l=1}^{L=4} = FPN(Swin(F_{RD}^{fuse})) 
\end{equation}
\begin{equation}
    \left \{ F_{DZ}^{l}  \right \}_{l=1}^{L=4} = FPN(Swin(F_{DZ}^{fuse})) 
\end{equation}
\begin{equation}
    \left \{ F_{ZR}^{l}  \right \}_{l=1}^{L=4} = FPN(Swin(F_{ZR}^{fuse})) 
\end{equation}
Furthermore, since we know each scale voxel grid Cartesian coordinate of the final 3D occupancy volume in 3D space, we leverage Cartesian to polar coordinate conversion and the bilinear interpolate to obtain the final four scales 3D feature volumes under Cartesian coordinate. This process is described as:
\begin{equation}
    \left \{ V_{Cartesian}^{l} \right \}_{l=1}^{L=4} = Bilinear(c2d(\left \{ x_{i}^{l},y_{i}^{l},z_{i}^{l} \right \} )) 
\end{equation}
where $V_{Cartesian}^{l}$ refers to the $l$-th scale 3D feature volume under Cartesian coordinates, $c2d$ refers to the Cartesian to cylindrical coordinates conversion, $\left \{ x_{i}^{l},y_{i}^{l},z_{i}^{l} \right \}$ refers to each $i$-th scale 3D feature volume's voxel grid Cartesian coordinates. Lastly, the multi-scale supervision is leveraged on four-scale 3D feature volumes to train the model.

\section{Experimental Results}\label{simulation}
\subsection{Implementation Details}
The OccCylindrical utilises ResNet50 \cite{residual, deformable} as the 2D backbone, with pre-trained weights provided by FCOS3D \cite{fcos3d}, to extract image features. The feature maps of the backbone from stages 0, 1, 2, and 3 are fed into FPN \cite{fpn}, resulting in a single visual feature. Our approach uses 10 LiDAR sweeps for each data sample. AdamW Optimiser with an initial learning rate of 5e-5 and weight decay of 0.01 is used for optimisation. The learning rate is decayed using a multistep scheduler. The model is trained on three A40 GPUs, each with 48GB of memory, for two days.

\subsection{Loss Function}
The model is trained with focal loss \cite{focal} ($L_{focal}$), Lovasz-softmax loss \cite{lovasz} ($L_{lovasz}$), and scene-class affinity loss \cite{monoscene} ($L_{scal_geo}$ and $L_{scal_sem}$). Considering the significance of high-resolution 3D volumes compared to lower-resolution ones, a decayed loss weight $w=\frac{1}{2^{l}} $ is applied for each resolution supervision. For depth supervision, a binary cross-entropy loss ($L_{bce}^{depth}$) is used to refine the depth distribution feature. The final loss is as follows:
\begin{equation}
    Occ_{loss} = \sum_{l=0}^{3} \frac{1}{2^{l} } \times (L_{focal}^{l} + L_{lovasz}^{l} + L_{scal\_geo}^{l} + L_{scal\_sem}^{l})
\end{equation}
\begin{equation}
    Loss = Occ_{loss} + \lambda L_{bce}^{depth}
\end{equation}
where $l$ refer to $l$-th 3D volume, and $\lambda$ balances the loss weight. In practice, we set $\lambda=3$. 

\subsection{Dataset}
Our 3D semantic occupancy prediction experiments were conducted using the nuScenes dataset, with ground truth labels obtained from SurroundOcc \cite{surroundocc}. We defined the range of the X and Y axes as [-50, 50] meters and the range of the Z axis as [-5, 3] meters in the LiDAR coordinates for the prediction of 3D semantic occupancy. The ground truth of each data sample has a resolution of $200\times200\times16$, with a voxel size of 0.5m. Due to the absence of annotations in the test set, we train our model on the training set and evaluate its performance on the validation set. Furthermore, following the methodology proposed in \cite{occfusion}, we conduct an in-depth analysis of the performance of our model in challenging scenarios, especially rainy and night conditions. 

\subsection{Performance Evaluate Metrics}
\begin{table*}[htpb]
\vspace{2mm}
  \centering
  \begin{adjustbox}{width=\textwidth}
  \begin{tabular}{c|c|c|cc|cccccccccccccccc}
    \toprule
    Method & Img Backbone & Modality & IoU & mIoU & \rotatebox{90}{\textcolor{barrier}{$\bullet$} barrier} & \rotatebox{90}{\textcolor{bicycle}{$\bullet$} bicycle} & \rotatebox{90}{\textcolor{bus}{$\bullet$} bus} & \rotatebox{90}{\textcolor{car}{$\bullet$} car} & \rotatebox{90}{\textcolor{construction}{$\bullet$} const. veh.} & \rotatebox{90}{\textcolor{motorcycle}{$\bullet$} motorcycle} & \rotatebox{90}{\textcolor{pedestrian}{$\bullet$} pedestrian} & \rotatebox{90}{\textcolor{cone}{$\bullet$} traffic cone} & \rotatebox{90}{\textcolor{trailer}{$\bullet$} trailer} & \rotatebox{90}{\textcolor{truck}{$\bullet$} truck} & \rotatebox{90}{\textcolor{driveable}{$\bullet$} drive. surf.} & \rotatebox{90}{\textcolor{flat}{$\bullet$} other flat} & \rotatebox{90}{\textcolor{sidewalk}{$\bullet$} sidewalk} & \rotatebox{90}{\textcolor{terrain}{$\bullet$} terrain} & \rotatebox{90}{\textcolor{manmade}{$\bullet$} manmade} & \rotatebox{90}{\textcolor{vegetation}{$\bullet$} vegetation} \\
     \midrule 
    MonoScene \cite{monoscene} & R101-DCN & C & 23.96 & 7.31 & 4.03 & 0.35 & 8.00 & 8.04 & 2.90 & 0.28 & 1.16 & 0.67 & 4.01 & 4.35 & 27.72 & 5.20 & 15.13 & 11.29 & 9.03 & 14.86\\
    Atlas \cite{atlas} & R101-DCN & C & 28.66 & 15.00 & 10.64 & 5.68 & 19.66 & 24.94 & 8.90 & 8.84 & 6.47 & 3.28 & 10.42 & 16.21 & 34.86 & 15.46 & 21.89 & 20.95 & 11.21 & 20.54 \\
    BEVFormer \cite{bevformer} & R101-DCN & C & 30.50 & 16.75 & 14.22 & 6.58 & 23.46 & 28.28 & 8.66 & 10.77 & 6.64 & 4.05 & 11.20 & 17.78 & 37.28 & 18.00 & 22.88 & 22.17 & 13.80 & 22.21 \\
    TPVFormer \cite{tpvformer} & R101-DCN & C & 30.86 & 17.10 & 15.96 & 5.31 & 23.86 & 27.32 & 9.79 & 8.74 & 7.09 & 5.20 & 10.97 & 19.22 & 38.87 & 21.25 & 24.26 & 23.15 & 11.73 & 20.81 \\
    C-CONet \cite{openoccupancy} & Res101 & C & 26.10 & 18.40 & 18.60 & 10.00 & 26.40 & 27.40 & 8.60 & 15.70 & 13.30 & 9.70 & 10.90 & 20.20 & 33.00 & 20.70 & 21.40 & 21.80 & 14.70 & 21.30 \\
    Inv.MatrixVT3D \cite{inversematrixvt3d} & R101-DCN & C & 30.03 & 18.88 & 18.39 & 12.46 & 26.30 & 29.11 & 11.00 & 15.74 & 14.78 & 11.38 & 13.31 & 21.61 & 36.30 & 19.97 & 21.26 & 20.43 & 11.49 & 18.47 \\
    RenderOcc \cite{renderocc} & Res101 & C & 29.20 & 19.00 & 19.70 & 11.20 & 28.10 & 28.20 & 9.80 & 14.70 & 11.80 & 11.90 & 13.10 & 20.10 & 33.20 & 21.30 & 22.60 & 22.30 & 15.30 & 20.90 \\
    SurroundOcc \cite{surroundocc} & R101-DCN & C & 31.49 & 20.30 & 20.59 & 11.68 & 28.06 & 30.86 & 10.70 & 15.14 & 14.09 & 12.06 & 14.38 & 22.26 & 37.29 & 23.70 & 24.49 & 22.77 & 14.89 & 21.86 \\
    LMSCNet \cite{roldao2020lmscnet} & - & L & 36.60 & 14.90 & 13.10 & 4.50 & 14.70 & 22.10 & 12.60 & 4.20 & 7.20 & 7.10 & 12.20 & 11.50 & 26.30 & 14.30 & 21.10 & 15.20 & 18.50 & 34.20 \\
    L-CONet \cite{openoccupancy} & - & L & 39.40 & 17.70 & 19.20 & 4.00 & 15.10 & 26.90 & 6.20 & 3.80 & 6.80 & 6.00 & 14.10 & 13.10 & 39.70 & 19.10 & 24.00 & 23.90 & 25.10 & 35.70 \\
    M-CONet \cite{openoccupancy} & Res101 & C+L & 39.20 & 24.70 & 24.80 & 13.00 & 31.60 & 34.80 & 14.60 & 18.00 & 20.00 & 14.70 & 20.00 & 26.60 & 39.20 & 22.80 & 26.10 & 26.00 & 26.00 & 37.10 \\
    Co-Occ \cite{pan2024co} & Res101 & C+L & 41.10 & 27.10 & 28.10 & 16.10 & 34.00 & \textbf{37.20} & 17.00 & 21.60 & 20.80 & 15.90 & 21.90 & 28.70 & 42.30 & \textbf{25.40} & \textbf{29.10} & \textbf{28.60} & 28.20 & 38.00 \\
    OccFusion \cite{occfusion} & R101-DCN & C+R & 32.90 & 20.73 & 20.46 & 13.98 & 27.99 & 31.52 & 13.68 & 18.45 & 15.79 & 13.05 & 13.94 & 23.84 & 37.85 & 19.60 & 22.41 & 21.20 & 16.16 & 21.81 \\
    OccFusion \cite{occfusion} & R101-DCN & C+L & 43.53 & 27.55 & 25.15 & 19.87 & 34.75 & 36.21 & 20.03 & 23.11 & 25.25 & 17.50 & \textbf{22.70} & 30.06 & 39.47 & 23.26 & 25.68 & 27.57 & 29.54 & 40.60 \\
    OccFusion \cite{occfusion} & R101-DCN & C+L+R & 43.96 & 28.27 & \textbf{28.32} & 20.95 & \textbf{35.06} & 36.84 & \textbf{20.33} & 26.22 & 25.86 & 19.17 & 21.27 & \textbf{30.64} & 40.08 & 23.75 & 25.56 & 27.63 & 29.82 & 40.82 \\
    \hline
    \begin{tabular}[c]{@{}c@{}}OccCylindrical\\ (Ours)\end{tabular} & ResNet50 & C+L & \textbf{44.94} & \textbf{28.67} & 26.17 & \textbf{22.12} & 31.47 & 36.84 & 17.95 & \textbf{27.77} & \textbf{29.85} & \textbf{23.90} & 20.64 & 28.27 & \textbf{43.00} & 23.14 & 27.99 & 27.81 & \textbf{30.81} & \textbf{40.95} \\
    \bottomrule
  \end{tabular}
  \end{adjustbox}
  \caption{\small \textbf{3D semantic occupancy prediction results on SurroundOcc-nuScenes validation set}. All methods are trained with dense occupancy labels from SurroundOcc \cite{surroundocc}. Notion of modality: Camera (C), LiDAR (L), Radar (R).}
  \label{occ}
\end{table*}

\begin{table*}[htpb]
  \centering
  \begin{adjustbox}{width=\textwidth}
  \begin{tabular}{c|c|c|cc|cccccccccccccccc}
    \toprule
    Method & Backbone & Modality & IoU & mIoU & \rotatebox{90}{\textcolor{barrier}{$\bullet$} barrier} & \rotatebox{90}{\textcolor{bicycle}{$\bullet$} bicycle} & \rotatebox{90}{\textcolor{bus}{$\bullet$} bus} & \rotatebox{90}{\textcolor{car}{$\bullet$} car} & \rotatebox{90}{\textcolor{construction}{$\bullet$} const. veh.} & \rotatebox{90}{\textcolor{motorcycle}{$\bullet$} motorcycle} & \rotatebox{90}{\textcolor{pedestrian}{$\bullet$} pedestrian} & \rotatebox{90}{\textcolor{cone}{$\bullet$} traffic cone} & \rotatebox{90}{\textcolor{trailer}{$\bullet$} trailer} & \rotatebox{90}{\textcolor{truck}{$\bullet$} truck} & \rotatebox{90}{\textcolor{driveable}{$\bullet$} drive. surf.} & \rotatebox{90}{\textcolor{flat}{$\bullet$} other flat} & \rotatebox{90}{\textcolor{sidewalk}{$\bullet$} sidewalk} & \rotatebox{90}{\textcolor{terrain}{$\bullet$} terrain} & \rotatebox{90}{\textcolor{manmade}{$\bullet$} manmade} & \rotatebox{90}{\textcolor{vegetation}{$\bullet$} vegetation} \\
     \midrule 
    Co-Occ \cite{pan2024co} & R101 & C+L & 40.30 & 26.60 & 26.60 & 19.10 & \textbf{37.60} & \textbf{37.20} & 15.90 & 20.30 & 16.30 & 12.30 & 23.30 & 27.00 & 41.00 & 22.80 & \textbf{35.20} & \textbf{24.60} & 27.80 & 39.30 \\
    OccFusion \cite{occfusion} & R101-DCN & C+L & 42.67 & 26.68 & 20.91 & 18.39 & 35.26 & 36.19 & \textbf{17.69} & 19.05 & 19.40 & 17.08 & \textbf{23.86} & \textbf{28.86} & 38.28 & \textbf{26.37} & 31.44 & 21.35 & 29.48 & 43.22 \\
    OccFusion \cite{occfusion} & R101-DCN & C+L+R & 42.67 & 27.39 & \textbf{27.82} & 21.10 & 36.00 & 37.10 & 17.23 & 21.67 & 20.34 & 17.46 & 20.93 & 28.57 & 38.99 & 24.72 & 31.96 & 21.26 & 29.64 & \textbf{43.53} \\
    \hline
    \begin{tabular}[c]{@{}c@{}}OccCylindrical\\ (Ours)\end{tabular} & ResNet50 & C+L & \textbf{44.08} & \textbf{28.07} & 24.50 & \textbf{22.66} & 33.47 & 36.79 & 16.96 & \textbf{23.63} & \textbf{23.79} & \textbf{23.49} & 22.70 & 28.14 & \textbf{41.67} & 22.32 & 33.55 & 21.44 & \textbf{30.58} & 43.45 \\
    \bottomrule
  \end{tabular}
  \end{adjustbox}
  \caption{\small \textbf{3D semantic occupancy prediction results on SurroundOcc-nuScenes validation rainy scenario subset}. All methods are trained with dense occupancy labels from \cite{surroundocc}. Notion of modality: Camera (C), LiDAR (L), Radar (R).}
  \label{occ_rainy}
\end{table*}

\begin{table*}[htpb]
  \centering
  \begin{adjustbox}{width=\textwidth}
  \begin{tabular}{c|c|c|cc|cccccccccccccccc}
    \toprule
    Method & Backbone & Modality & IoU & mIoU & \rotatebox{90}{\textcolor{barrier}{$\bullet$} barrier} & \rotatebox{90}{\textcolor{bicycle}{$\bullet$} bicycle} & \rotatebox{90}{\textcolor{bus}{$\bullet$} bus} & \rotatebox{90}{\textcolor{car}{$\bullet$} car} & \rotatebox{90}{\textcolor{construction}{$\bullet$} const. veh.} & \rotatebox{90}{\textcolor{motorcycle}{$\bullet$} motorcycle} & \rotatebox{90}{\textcolor{pedestrian}{$\bullet$} pedestrian} & \rotatebox{90}{\textcolor{cone}{$\bullet$} traffic cone} & \rotatebox{90}{\textcolor{trailer}{$\bullet$} trailer} & \rotatebox{90}{\textcolor{truck}{$\bullet$} truck} & \rotatebox{90}{\textcolor{driveable}{$\bullet$} drive. surf.} & \rotatebox{90}{\textcolor{flat}{$\bullet$} other flat} & \rotatebox{90}{\textcolor{sidewalk}{$\bullet$} sidewalk} & \rotatebox{90}{\textcolor{terrain}{$\bullet$} terrain} & \rotatebox{90}{\textcolor{manmade}{$\bullet$} manmade} & \rotatebox{90}{\textcolor{vegetation}{$\bullet$} vegetation} \\
     \midrule 
    Co-Occ \cite{pan2024co} & R101 & C+L & 35.60 & 14.60 & 8.40 & 16.40 & 0.00 & 37.20 & 0.00 & 13.80 & 10.90 & 0.40 & 0.00 & 24.30 & 36.40 & 2.20 & 14.60 & 17.00 & 19.90 & 31.70 \\
    OccFusion \cite{occfusion} & R101-DCN & C+L & 40.87 & 15.87 & 13.28 & \textbf{17.53} & 0.00 & 36.42 & 0.00 & 16.16 & 11.37 & \textbf{1.42} & 0.00 & 25.71 & 32.64 & 0.63 & 16.06 & 20.87 & 24.52 & 37.27 \\
    OccFusion \cite{occfusion} & R101-DCN & C+L+R & 41.01 & 16.61 & 15.70 & 16.26 & 0.00 & \textbf{38.09} & 0.00 & 22.18 & 13.24 & 0.08 & 0.00 & 25.92 & 33.15 & 1.57 & 16.08 & 21.09 & 24.51 & 37.83 \\
    \hline \begin{tabular}[c]{@{}c@{}}OccCylindrical\\ (Ours)\end{tabular}
     & ResNet50 & C+L & \textbf{43.38} & \textbf{17.79} & \textbf{16.19} & 10.04 & 0.00 & 37.84 & 0.00 & \textbf{25.63} & \textbf{13.28} & 0.23 & 0.00 & \textbf{32.93} & \textbf{39.59} & \textbf{3.89} & \textbf{17.89} & \textbf{21.61} & \textbf{26.99} & \textbf{38.47} \\
    \bottomrule
  \end{tabular}
  \end{adjustbox}
  \caption{\small \textbf{3D semantic occupancy prediction results on SurroundOcc-nuScenes validation night scenario subset}. All methods are trained with dense occupancy labels from \cite{surroundocc}. Notion of modality: Camera (C), LiDAR (L), Radar (R).}
  \label{occ_night}
\end{table*}

To assess the performance of various SOTA algorithms and compare them with our approach in the 3D semantic occupancy prediction task, we use intersection over union (IoU) to evaluate each semantic class. Moreover, we employ the mean IoU (mIoU) over all semantic classes as a comprehensive evaluation metric:
\begin{equation}
    IoU=\frac{TP}{TP+FP+FN} 
\end{equation}
and
\begin{equation}
    mIoU=\frac{1}{Cls}\sum_{i=1}^{Cls}  \frac{TP_{i}}{TP_{i}+FP_{i}+FN_{i}} 
\end{equation}
where $TP$, $FP$, and $FN$ represent the counts of true positives, false positives, and false negatives in our predictions, respectively, while $Cls$ denotes the total class number.

\subsection{Model Performance Analysis}

We evaluated the performance of our approach, OccClindrical, by comparing it with SOTA algorithms and presenting the results in Table \ref{occ}. Our method outperforms existing single 3D supervision signal-trained multi-modality SOTA methods ranking first on the benchmark. Our dual-modality approach outperforms OccFusion (C+L+R), a three-sensor fusion method, despite using fewer modalities. Our model excels at detecting small dynamic objects, such as bicycles, motorcycles, and pedestrians, whose identification is vital for AVs operation.

In addition, to fully evaluate the capabilities of our model in challenging conditions such as rainy and night scenarios, we use the annotation proposed in \cite{occfusion} for model evaluation. We perform a comparative analysis of our approach against other SOTA multimodality fusion approaches trained with single 3D supervision signals in terms of performance in rainy and night scenarios, which is detailed in Table \ref{occ_rainy} and Table \ref{occ_night}. In rainy scenarios, all algorithms exhibit varying degrees of performance decline, and our approach experiences a performance degradation of 0.86\% under the IoU metric and a performance degradation of 0.6\% under the mIoU metric. Despite this, our approach still ranks first in the benchmark and maintains small dynamic object detection capability, showcasing superior robustness compared to other SOTA approaches in rainy scenarios. In nighttime scenarios, all algorithms show reduced IoU and significantly lower mIoU performance due to limited semantic data from the camera. Despite this, our method ranks highest, demonstrating stronger resilience than other SOTA approaches in low-light conditions.

\subsection{Qualitative Study in Challenging Scenarios}
\begin{figure*}[htbp]
\centering
\begin{subfigure}[]{0.95\textwidth}
\includegraphics[width=0.95\textwidth]{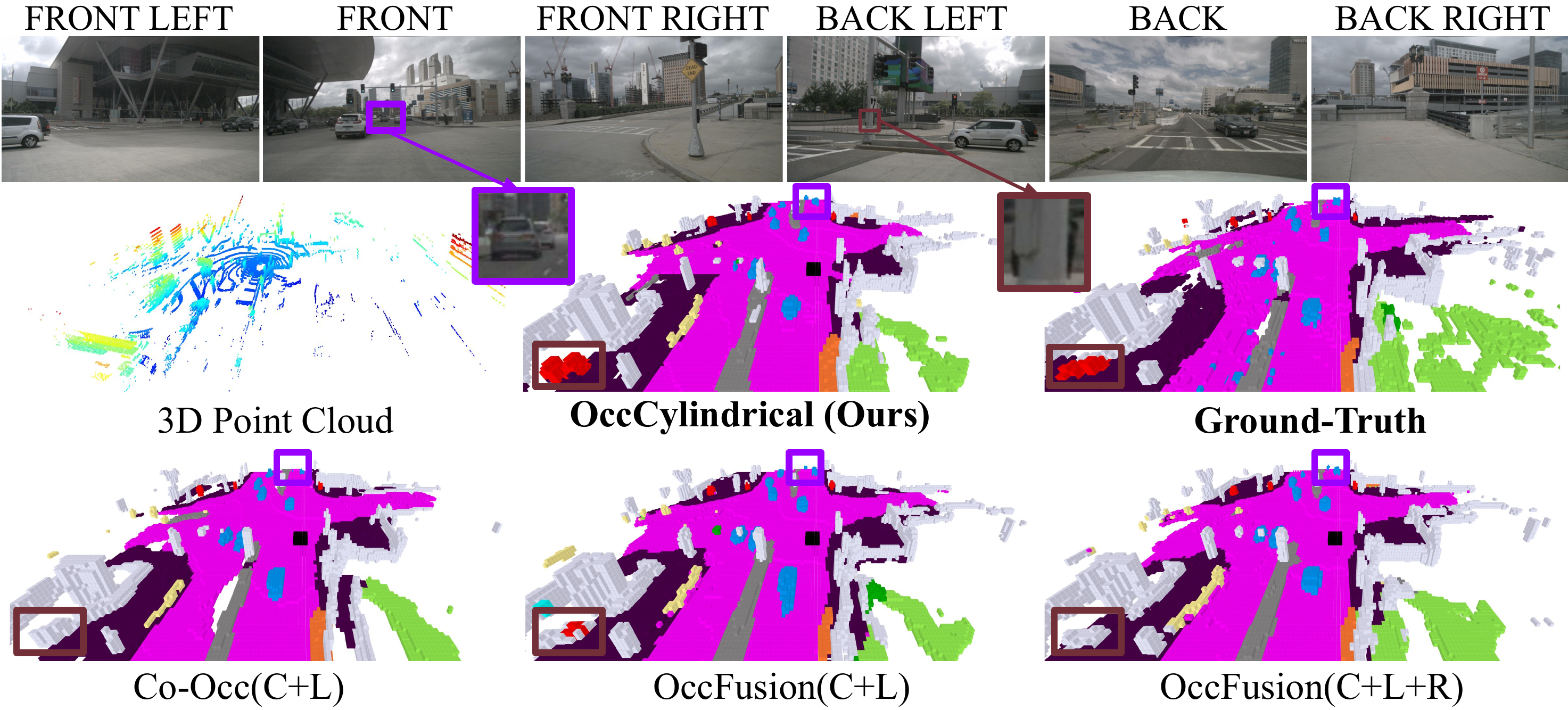}\\
\includegraphics[width=0.95\textwidth]{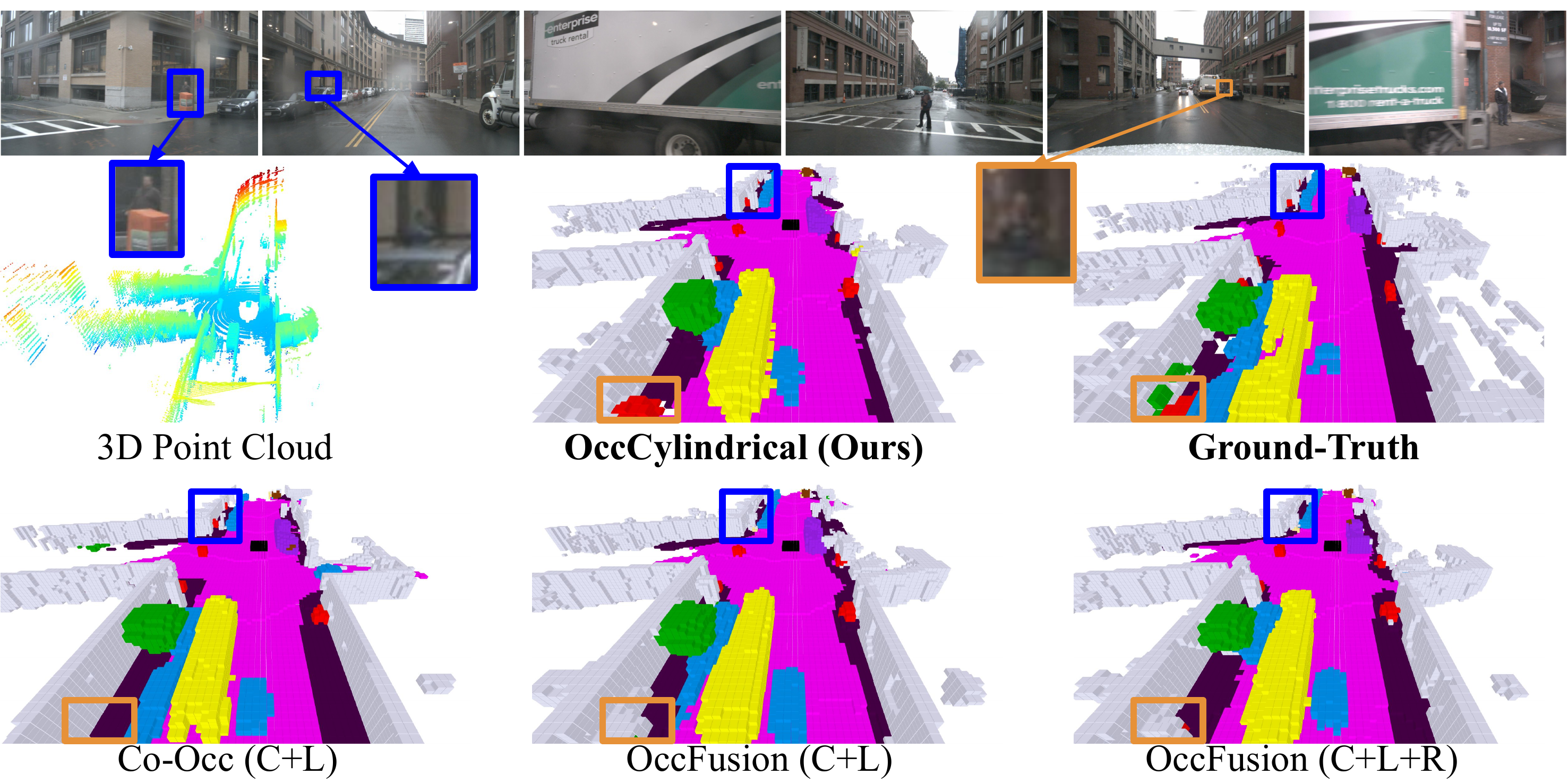}\\
\includegraphics[width=0.95\textwidth]{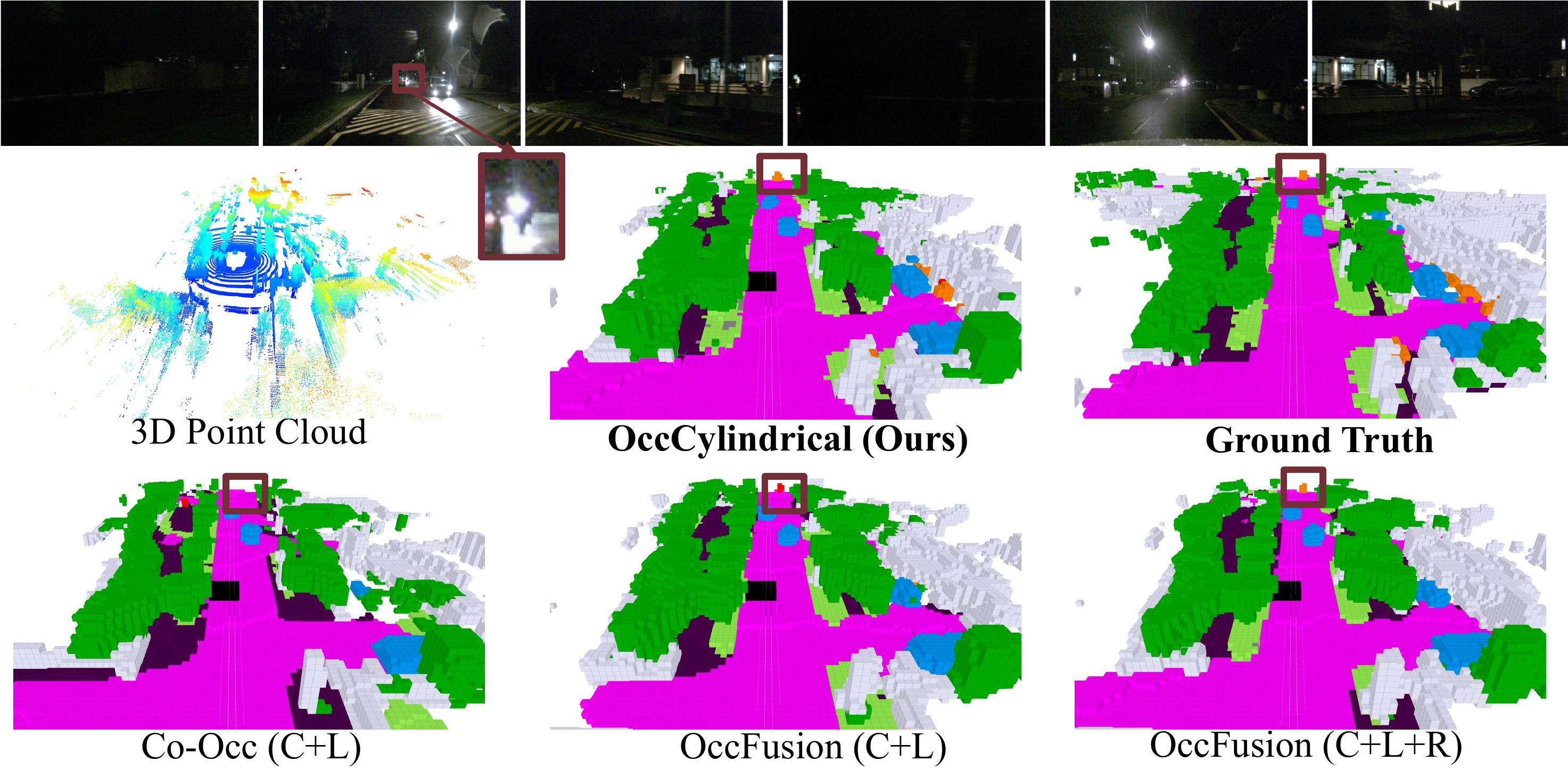}
\end{subfigure}
\caption{Qualitative study results for daytime, rainy, and nighttime scenarios displayed in the upper, middle, and bottom.}
\label{qualitative}
\end{figure*}
The qualitative study compares our approach's visualisation results against other SOTA algorithms under challenging scenarios, including rainy and nighttime conditions. The visualisation results are shown in Figure \ref{qualitative}. In the daytime scenario, shown at the top of Figure \ref{qualitative}, our algorithm successfully detects both a moving vehicle and a pedestrian in the distant region, as highlighted by the purple and dark red boxes. In particular, it accurately identifies partially occluded pedestrians behind a pole, demonstrating a superior ability to preserve fine-grained geometric and semantic information. Similarly, in the rainy scenario, as shown in the middle of Figure \ref{qualitative}, our algorithm is the only approach that successfully detects all partially occluded pedestrians in distant areas. In the night-time scenario, as shown at the bottom of Figure \ref{qualitative}, our algorithm is only based on two modalities that successfully detect the motorcycle in the distant area and achieve similar detection results to OccFusion (C + L + R), which uses three modalities.

\subsection{Model Efficiency Study}
\begin{table}[htbp]
  \centering
  \begin{adjustbox}{width=0.98\columnwidth}
  \begin{tabular}{c|c|ccc}
    \toprule
    {Method} & {Modality} & {\begin{tabular}[c]{@{}c@{}}Latency\\ (ms) ($\downarrow$)\end{tabular}} & {  \begin{tabular}[c]{@{}c@{}}Memory\\ (GB) ($\downarrow$)\end{tabular}} & Params \\
    \midrule
    SurroundOcc \cite{surroundocc} & C & 472 & 5.98 & 180.51M \\
    InverseMatrixVT3D \cite{inversematrixvt3d} & C & 447 & \textbf{4.41} & 67.18M \\
    OccFusion(C+R) \cite{occfusion} & C+R & 588 & 5.56 & 92.71M \\
    OccFusion(C+L) \cite{occfusion} & C+L & 591 & 5.56 & 92.71M \\
    OccFusion(C+L+R) \cite{occfusion} & C+L+R & 601 & 5.78 & 114.97M \\
    \hline
    OccCylindrical (Ours) & C+L & \textbf{398} & 10.63 & 111.62M \\
    \bottomrule
  \end{tabular}
  \end{adjustbox}
  \caption{\small Model efficiency comparison of different methods. The experiments are performed on a single A10 using six multi-camera images, LiDAR, and radar data. For input image resolution, all methods adopt $1600\times900$. $\downarrow$:the lower, the better.}
  \label{efficiency}
\end{table}
We evaluated the efficiency of our proposed approach and demonstrated its performance against other SOTA algorithms in Table \ref{efficiency}. Since we use ResNet50 as the 2D backbone, which is smaller than the ResNet101DCN that other SOTA algorithms use, and we eliminate all Conv3D-based operations by compressing Cylindrical volume into three TPV-Polar planes and densely use Conv2D-based operations to refine the feature, our approach runs faster than other SOTA algorithms. However, due to the high density of the pseudo-3D point cloud, GPU memory consumption is around 5GB higher than that of other SOTA algorithms.

\subsection{Ablation Study Of Model Component}
\subsubsection{Ablation study on depth-estimation module}
We investigate the influence of each component of the depth estimation module on the overall performance impact and exhibit the results of the experiment in Table \ref{Depth}.
\begin{table}[htbp]
  \centering
  \begin{adjustbox}{width=\columnwidth}
  \begin{tabular}{ccc|cc}
    \toprule
    \begin{tabular}[c]{@{}c@{}}Depth\\ Supervision\end{tabular} & \begin{tabular}[c]{@{}c@{}}Depth\\ Feature\end{tabular} & \begin{tabular}[c]{@{}c@{}}Positional\\ Embedding\end{tabular} & mIoU ($\uparrow$) & IoU ($\uparrow$) \\
    \midrule
    \Checkmark & \Checkmark & \Checkmark & 40.71\% (\textcolor{green}{Baseline}) & 22.08\% (\textcolor{green}{Baseline}) \\
    \Checkmark & \Checkmark & \XSolidBrush & 39.04\% (\textcolor{red}{-1.67\%}) & 19.61\% (\textcolor{red}{-2.47\%}) \\ 
    \Checkmark & \XSolidBrush & \Checkmark & 36.58\% (\textcolor{red}{-4.13\%}) & 19.33\% (\textcolor{red}{-2.75\%}) \\
    \XSolidBrush & \Checkmark & \Checkmark & 38.61\% (\textcolor{red}{-2.10\%}) & 19.00\% (\textcolor{red}{-3.08\%}) \\
    \bottomrule
  \end{tabular}
  \end{adjustbox}
  \caption{\small Ablation study results on the depth-estimation module used in the framework. Depth supervision from LiDAR 3D point cloud, Fepth feature, the output of depthnet and Positional embedding, the output of depthnet. $\uparrow$:the higher, the better.}
  \label{Depth}
\end{table}
Each component impacts the depth estimation module, with positional embedding contributing the least, followed by depth distribution features and depth supervision.

\subsubsection{Ablation study on the density of pseudo-3D point cloud}
The ablation study on pseudo-3D point cloud density and its impact on model performance is summarized in Table \ref{density}.
\begin{table}[htbp]
  \centering
  \begin{adjustbox}{width=\columnwidth}
  \begin{tabular}{c|cccc}
    \toprule
    \begin{tabular}[c]{@{}c@{}}Interval\\ $d$\end{tabular} & Density & Memory ($\downarrow$) & mIoU ($\uparrow$) & IoU ($\uparrow$) \\
    \midrule
    1 m & 1740000 / 552178 & 10.63 GB & 22.08\% (\textcolor{green}{Baseline}) & 40.71\% (\textcolor{green}{Baseline}) \\
    2 m & 870000 / 279359 & 6.77 GB & 17.27\% (\textcolor{red}{-4.81\%}) & 38.41\% (\textcolor{red}{-2.30\%}) \\ 
    10 m & 174000 / 57612 & 5.50 GB & 15.96\% (\textcolor{red}{-6.12\%}) & 36.03\% (\textcolor{red}{-4.68\%}) \\
    \bottomrule
  \end{tabular}
  \end{adjustbox}
  \caption{\small Ablation study results on the density of pseudo-3D point cloud. Interval $d$ refers to the depth bin interval along each depth-estimation ray of each visual feature grid. The density column shows the total pseudo-3D points before and after the max pooling operation. $\uparrow$: the higher, the better. $\downarrow$: the lower, the better.}
  \label{density}
\end{table}
We adjusted the density of the pseudo-3D point cloud by setting different depth-bin interval values. The small interval results in a much denser pseudo-3D point cloud that gives better model performance; however, denser pseudo-3D point clouds consume more GPU memory during the inference time, as shown in the Table \ref{density}.

\subsubsection{Ablation study on TPV-Polar-Fusion}
The ablation study regarding the TPV-Polar Fusion is performed by setting different group parameter values $M$ and evaluating each set's impact on overall model performance. The experimental results are shown in Table \ref{M}.
\begin{table}[htbp]
  \centering
  \begin{adjustbox}{width=0.75\columnwidth}
  \begin{tabular}{c|cc}
    \toprule
    M Group & mIoU ($\uparrow$) & IoU ($\uparrow$) \\
    \midrule
    8 & 41.94\% (\textcolor{green}{Baseline}) & 23.66\% (\textcolor{green}{Baseline}) \\
    4 & 41.13\% (\textcolor{red}{-0.81\%}) & 22.52\% (\textcolor{red}{-1.14\%}) \\ 
    2 & 41.55\% (\textcolor{red}{-0.39\%}) & 22.43\% (\textcolor{red}{-1.23\%}) \\
    1 & 40.24\% (\textcolor{red}{-1.70\%}) & 21.33\% (\textcolor{red}{-2.33\%}) \\
    \bottomrule
  \end{tabular}
  \end{adjustbox}
  \caption{\small Ablation study results on TPV-Polar Fusion module. We adjust the group parameter setting $M$ in the spatial group pooling module to study its impact on the model's overall performance. $\uparrow$: the higher, the better.}
  \label{M}
\end{table}
As the total number of groups decreases by setting the smaller $M$ value, the model experiences performance degradation due to the loss of more fine-grained geometry information. 

\section{Conclusion}\label{conclusion}
This study presents OccCylindrical, a novel multisensor framework that integrates surround-view cameras and LiDAR to perform the 3D semantic occupancy prediction. Our proposed method leverages a depth-estimation module together with the LiDAR's 3D point cloud supervision to generate a high-quality pseudo-3D point cloud. Then, real and pseudo-3D point clouds are converted to cylindrical coordinates to better preserve fine-grained geometry information. A TPV-Polar-Fusion module is further proposed to compress and fuse the two cylindrical volumes into a set of TPV-Polar planes using spatial group pooling and 2D dynamic fusion, followed by a shared encoder-decoder module implemented in the 2D image backbone to refine the three TPV-Polar planes further. The experiments conducted on the nuScenes dataset, including challenging rainy and night-time scenarios, demonstrate the effectiveness and robustness of our method and preserve the distinct strength of each sensor.

\bibliographystyle{IEEEtran}
\bibliography{mybibtex}

\end{document}